\documentclass[letterpaper, 10 pt, conference]{ieeeconf}  

\IEEEoverridecommandlockouts                              

\usepackage{url}
\usepackage{graphicx}
\usepackage{amsmath}
\usepackage{amssymb}
\usepackage{booktabs}
\usepackage{balance}
\usepackage{multirow}
\usepackage{xcolor}
\usepackage{tabularx}
\usepackage{subcaption}
\usepackage{tikz}   

\title{\LARGE \bf
PASSAGE: Scaling Scene-Aligned Motion Learning for Perceptive Humanoid Traversal in Cluttered Environments
}

\author{%
Yuxuan Ma$^{1,*}$,
Zicheng Zeng$^{1,2,3,*}$,
Chunlin Peng$^{1,4,5,*}$,
Zhoujian Li$^{1,6}$,\\
Zetong Zhao$^{1,7}$,
Zhikai Zhang$^{1,7}$,
Yunrui Lian$^{1,7}$,
Han Xue$^{1,7}$,
Sikai Liang$^{1,7}$,\\
Weiyi Zhu$^{1}$,
Mulin Chen$^{1,7}$,
Chenghuai Lin$^{1}$,
Jiayu Zeng$^{1}$,
Yanwei An$^{1}$,\\
Songan Zhang$^{5}$,
Jiayuan Gu$^{3}$,
Jilong Wang$^{1}$,
Jingbo Wang$^{1}$,
He Wang$^{1,8}$,
and Li Yi$^{1,2,7,\dagger}$\\
{\normalsize
$^{1}$Galbot \quad
$^{2}$Shanghai Qi Zhi Institute \quad
$^{3}$ShanghaiTech University \quad
$^{4}$Zhongguancun Academy}\\
{\normalsize
$^{5}$Shanghai Jiao Tong University \quad
$^{6}$National University of Singapore \quad
$^{7}$Tsinghua University \quad
$^{8}$Peking University}\\
{\normalsize
$^{*}$Equal contribution
$^{\dagger}$Corresponding author}\\
{\small
Project page:
\url{https://galaxygeneralrobotics.github.io/PASSAGE/}}
}

\IEEEaftertitletext{%
    \vspace{0.4em}%
    \begin{center}%
    \begin{tikzpicture}[panel/.style={anchor=north west, xshift=3pt, yshift=-3pt,
            font=\bfseries\small, text=white, fill=black, fill opacity=0.55,
            text opacity=1, inner sep=2.5pt, rounded corners=1.5pt}]
        \node[anchor=south west, inner sep=0] (banner)
            {\includegraphics[width=\textwidth]{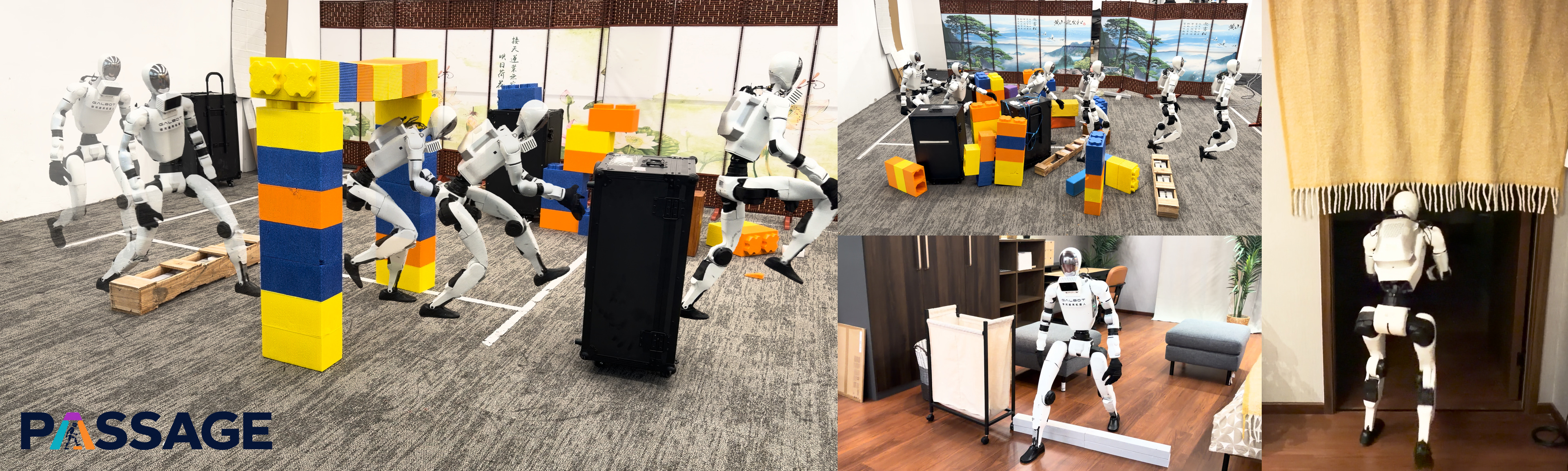}};
        \begin{scope}[x={(banner.south east)}, y={(banner.north west)}]
            \node[panel] at (0.000, 1.0) {(a)};
            \node[panel] at (0.534, 1.0) {(b)};
            \node[panel] at (0.534, 0.5) {(c)};
            \node[panel] at (0.804, 1.0) {(d)};
        \end{scope}
    \end{tikzpicture}%
    \captionof{figure}{PASSAGE enables a humanoid robot to traverse cluttered environments with whole-body behaviors using only onboard perception.
    \textbf{(a)}~Multi-exposure composite of the robot stepping over a wooden pallet, ducking under a block archway, and squeezing past a wheeled case.
    \textbf{(b)}~A cluttered test scene with the robot's poses overlaid along the traversed path.
    \textbf{(c)}~Stepping over a low barrier in a furnished living space.
    \textbf{(d)}~Ducking under a hanging curtain to pass through a doorway.
    All behaviors are produced by a single planner and tracker learned from scene-aligned human demonstrations, without prebuilt maps or offboard computation.}%
    \label{fig:teaser}%
    \end{center}%
    \vspace{0.2em}%
}

\begin{document}

\maketitle
\thispagestyle{empty}
\pagestyle{empty}

\begin{abstract}
Humanoid robots can step over, squeeze past, and duck under obstacles, but learning to select and coordinate these behaviors from onboard perception remains challenging. Many existing approaches rely on task-specific reinforcement-learning objectives or curated motion libraries, making broad behavioral coverage costly. We present PASSAGE, a perception-conditioned planner--tracker framework for humanoid traversal. Using virtual reality and inertial motion capture, we collect 100~h of scene-aligned human motion across 1{,}500 cluttered scenes. A conditional flow-matching planner generates short-horizon references from motion history, a local destination, and a robot-centric multi-layer elevation map, while a perceptive whole-body tracker executes them at 50~Hz with geometric feedback. Real-time chunking promotes inter-chunk consistency, and planner-side RL post-training under the frozen tracker further improves closed-loop performance. Without skill annotations or obstacle-specific policies, one planner--tracker pair selects and composes traversal behaviors across unseen geometries. In simulation, component ablations quantify the contribution of each stage. Across three independent training seeds, scaling captured data from 6 to 100~h increases mean contact-free success from 48.1\% to 68.9\% on held-out scenes, while the final model with validated scene augmentation reaches 70.3\%. The fully onboard system integrates egocentric 3D LiDAR perception, online occupancy mapping, 6.25-Hz planning, and 50-Hz control on a Jetson AGX Orin; tests across 50 unseen physical layouts demonstrate traversal without prebuilt maps or offboard computation.
\end{abstract}

\section{INTRODUCTION}

Humanoid robots can exploit whole-body articulation to traverse clutter by stepping over low obstacles, turning sideways through narrow openings, ducking under overhangs, or combining these behaviors. This requires a robot to select and coordinate motions from onboard geometry and execute them despite perception and tracking errors. Learning such a unified repertoire at scale remains challenging.

Recent perceptive methods acquire traversal behaviors primarily through reinforcement learning (RL). CAT~\cite{xue2026cat}, for example, trains scene-specific specialists with body-part-level collision guidance from HumanoidPF and distills them into a generalist. Although effective in clutter, this pipeline requires manually specified locomotion objectives and computationally intensive specialist training; its reported controller actuates only 12 leg joints, limiting the scope of whole-body coordination. Motion-data approaches~\cite{wu2026php,zhang2026learningwholebodyhumanoidlocomotion} provide richer whole-body priors, but scalable behavior selection in clutter requires demonstrations aligned with the surrounding geometry.

Wang et al.~\cite{wang2026moving} introduced a virtual-reality-based framework and collected 2.3~h of such demonstrations across 145 procedurally generated environments, focusing on data collection and benchmarking. These works leave two questions: how can scene-aligned demonstrations support a unified, closed-loop perception--planning--control system for diverse traversal, and how does demonstration scale affect motion generation and downstream execution?

We address both questions with PASSAGE, a scalable framework for perceptive whole-body humanoid traversal. Using a virtual-reality interface and inertial motion capture, we collect 100~h of scene-aligned human demonstrations across 1{,}500 procedurally generated cluttered scenes. A conditional flow-matching planner generates short-horizon references from motion history, a local destination, and robot-centric multi-layer elevation maps. Real-time chunking (RTC) promotes inter-chunk consistency, while planner-side RL post-training under the frozen tracker further improves closed-loop performance. A perception-enhanced ScaleBFM tracker~\cite{zeng2026scaling} executes the references using high-frequency geometric feedback. The same planner--tracker pair thereby selects and composes traversal behaviors across unseen geometries without skill annotations or obstacle-specific policies. We further quantify how captured scene-aligned data scale relates to closed-loop goal reaching and collision avoidance across three independent training seeds under a fixed planner architecture, tracker checkpoint, training budget, and evaluation protocol. Onboard 3D LiDAR and online occupancy mapping produce egocentric elevation observations, and the complete stack runs on an NVIDIA Jetson AGX Orin.

Our contributions are:
\begin{itemize}
    \item \textbf{Unified Perceptive Traversal.} We couple a flow-matching planner with a perception-enhanced whole-body tracker. RTC and planner-side RL post-training under a frozen tracker improve closed-loop execution, enabling the same pair to select, compose, and execute traversal behaviors without skill annotations or obstacle-specific policies.
    
    \item \textbf{Scene-Aligned Data Scaling.} Under a fixed architecture, training budget, tracker checkpoint, and evaluation protocol, we train planners with three independent seeds on each nested subset from 6 to 100~h and observe a robust positive empirical trend in closed-loop goal reaching and collision avoidance on held-out scenes.
    
    \item \textbf{Scalable Data Acquisition.} Our virtual-reality-guided inertial motion-capture pipeline combines procedural scene generation, retargeting, and MuJoCo-based kinematic collision validation, yielding 100~h of demonstrations across 1{,}500 cluttered scenes.
    
    \item \textbf{Fully Onboard Deployment.} We integrate egocentric 3D LiDAR perception, online mapping, planning, and control onboard the robot and validate traversal across 50 unseen physical layouts without prebuilt maps or offboard resources.
\end{itemize}

\section{RELATED WORK}
\subsection{Whole-Body Control and Perceptive Locomotion}

Following DeepMimic~\cite{2018-TOG-deepMimic}, generalist whole-body
trackers reproduce diverse references with one
policy~\cite{chen2025gmtgeneralmotiontracking,
luo2025sonicsupersizingmotiontracking,
zhao2025resmimicgeneralmotiontracking,exbody2,zhang2025any2track}.
They broaden the executable repertoire but neither choose future
motion nor ensure geometric compatibility.

Perceptive controllers condition execution on environmental
observations. KiVi~\cite{li2025kivi} separates proprioceptive and
visual pathways for robustness to visual corruption;
TAGA~\cite{li2026taga} learns terrain-aware attention; and Perceptive
BFM~\cite{wang2026perceptive} grounds prescribed references in local
terrain. For traversal, CAT~\cite{xue2026cat} distills
HumanoidPF-guided specialists through DAgger;
Gallant~\cite{ben2025gallantvoxelgridbasedhumanoid} and
PHP~\cite{wu2026php} learn goal-conditioned and skill-compositional
perceptive policies, respectively.
Light-Loco-Parkour~\cite{chen2026light} distills multiple
skills into one depth-conditioned policy that autonomously selects
behaviors without reference inputs, skill labels, or runtime motion
graphs. Prior systems thus learn behavior through simulator
objectives, compose curated skill libraries, or adapt supplied
references; PASSAGE instead learns destination-conditioned future
motion from long-horizon, scene-aligned demonstrations.

\subsection{Scene-Aligned Data and Motion Generation}

Scene-aligned data supervise geometry-conditioned behavior.
VideoMimic~\cite{allshire2025visual} reconstructs motion and geometry
from monocular video and distills an environment-conditioned
controller. EgoHTR~\cite{brandes2026egohtr} reconstructs
terrain-traversal demonstrations from egocentric wearables and
portable scans, validated with clip-specific perceptive trackers.
Moving Through Clutter~\cite{wang2026moving} collects 2.3~h across
145 procedural scenes as a data benchmark. PASSAGE scales virtual-reality-guided inertial capture to 100~h across
1{,}500 scenes and studies the empirical relationship between
captured-data scale and closed-loop traversal.

Generative models bridge navigation and whole-body control.
BeyondMimic~\cite{liao2026beyondmimic} guides diffusion at inference
using differentiable task costs. RLPF~\cite{yue2025rl} fine-tunes a
text-conditioned generator with tracker-derived physical feedback;
GenTrack~\cite{ling2026gentrack} alternates execution-grounded
generator alignment and tracker training. Zhang \emph{et
al.}~\cite{zhang2026learningwholebodyhumanoidlocomotion} train a
terrain-conditioned diffusion generator and an RL tracker separately,
then adapt the tracker in closed loop with the generator fixed.
PASSAGE instead pairs its destination-conditioned flow-matching
planner with a separately trained perceptive general tracker,
requiring no planner--tracker co-training; with the tracker frozen,
planner-side RL post-training further improves closed-loop traversal.

\begin{figure*}[t]
\centering
\includegraphics[width=\textwidth]{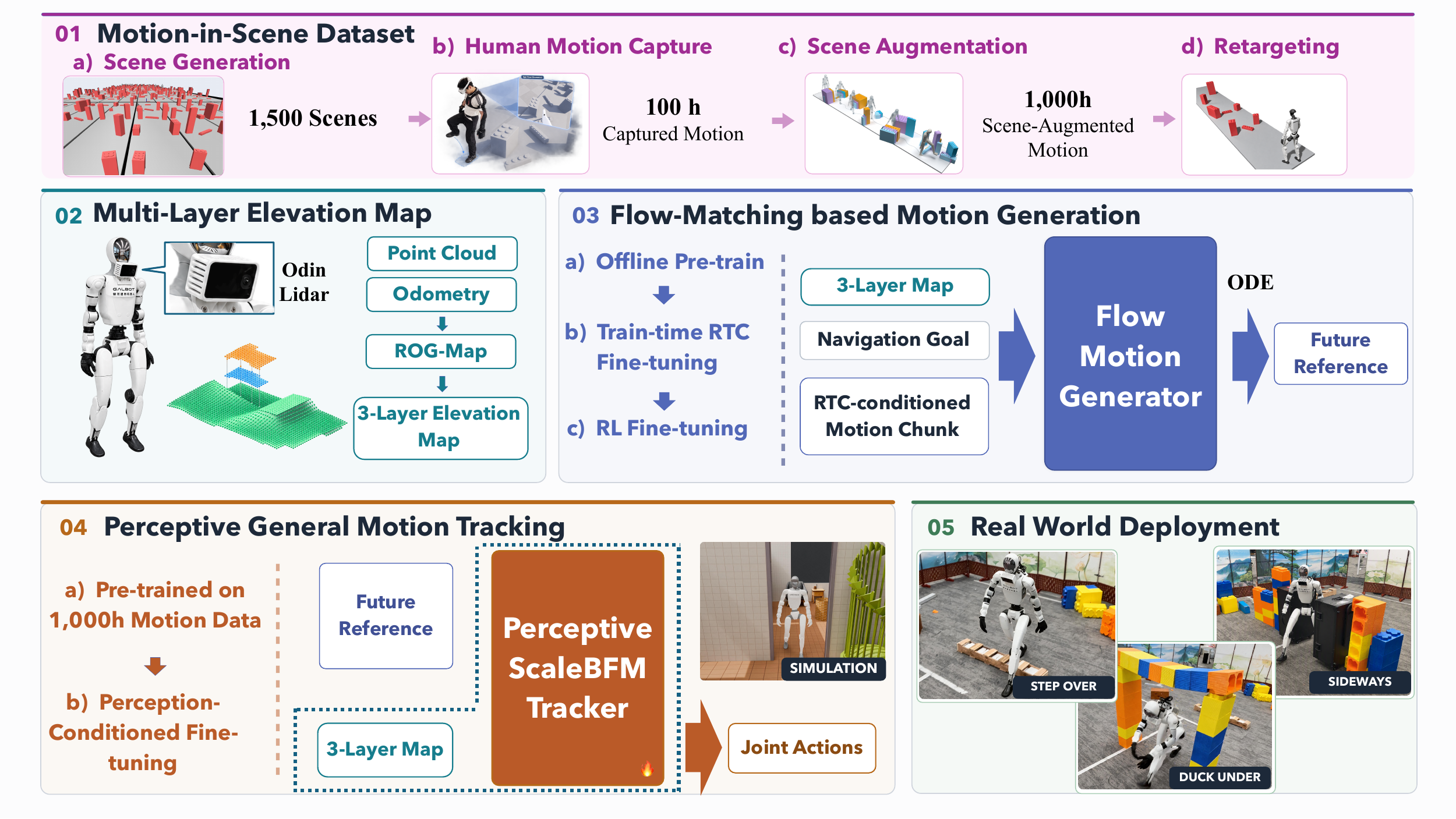}
\caption{Slow--fast plan-and-control architecture: the flow-matching planner (6.25\,Hz) outputs reference motions from the elevation map, and the perceptive controller (50\,Hz) executes them at the joint level.}
\label{fig:architecture}
\end{figure*}

\section{PASSAGE Framework}

PASSAGE separates geometry-conditioned motion generation from dynamic whole-body execution. At each replanning step, a destination-conditioned flow-matching planner generates a short-horizon kinematic reference from recent motion and robot-centric geometry. A perception-enhanced ScaleBFM tracker~\cite{zeng2026scaling} executes the reference at $50~\mathrm{Hz}$ through its general whole-body tracking interface. The planner and tracker are initially trained independently. The planner can roll out kinematic motion on its own; for dynamic execution, its outputs are deterministically converted into the reference representation consumed by the tracker. The tracker is trained with dataset references rather than planner outputs. During closed-loop post-training, the tracker remains frozen and only the planner is updated through tracker-executed rollouts. This design requires neither joint end-to-end optimization nor behavior-specific controllers.

\subsection{Scene-Aligned Demonstrations at Scale}
\label{sec:data}

\textbf{Procedural scenes.}
Using the procedural scene generator, we populate $10~\mathrm{m}$ corridors with composable block obstacles. Eight difficulty variables---$x$-step, lateral deviation, passage length and width, ceiling probability and height, and floor probability and height---are sampled and composed along each corridor to cover ground-level, lateral, and overhead constraints.

\begin{figure}[t]
\centering
\includegraphics[width=\columnwidth]{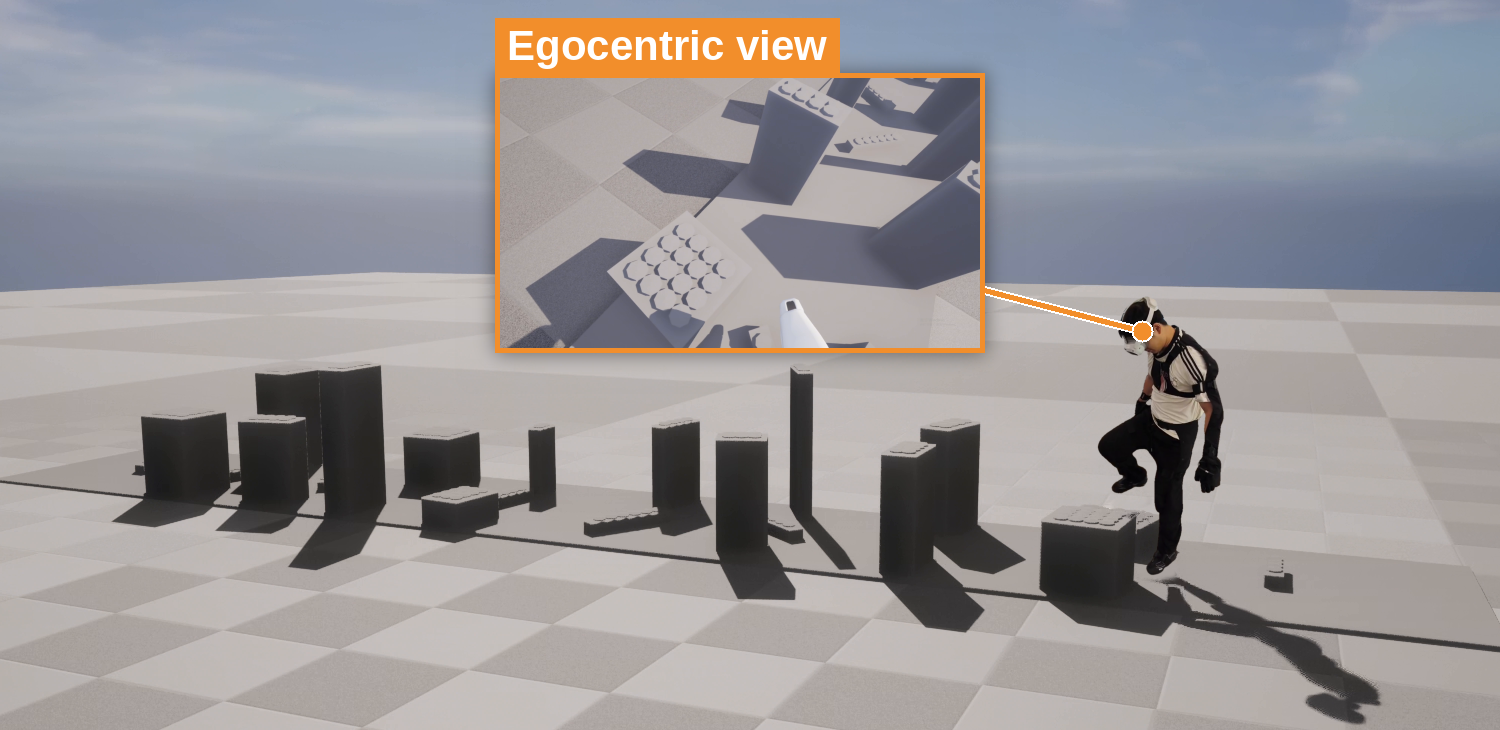}
\caption{VR-guided scene-aligned motion capture. An operator wearing an inertial motion-capture suit traverses a procedural scene from the avatar's egocentric view (inset).}
\label{fig:mocap}
\end{figure}

\textbf{VR-guided motion capture.}
Operators wearing a Noitom PN Link inertial motion-capture suit and a virtual reality (VR) headset traverse the generated scenes from an egocentric view (Fig.~\ref{fig:mocap}). Collisions between the virtual body and the scene trigger haptic feedback, and the corresponding demonstrations are rejected. Operators are encouraged to vary both their motions and traversal strategies. We retarget the recordings~\cite{omniretarget,araujo2025retargeting} to a 29-degree-of-freedom (DoF) Unitree G1 and scale the corresponding scenes to preserve motion--geometry alignment. After collision rejection and quality screening, we retain 19{,}310 sequences, 17{,}470{,}128 frames, totaling approximately $100~\mathrm{h}$ across 1{,}500 scenes.

\textbf{Validated scene augmentation.}
Since motion and scene geometry are stored separately, retargeting and
augmentation are independent and may be applied in either order. For
each motion, we generate ten obstacle variants with scales in
$[0.5,1.5]$ and rotations within $\pm15^\circ$, uniformly scale them
to match the retargeted motion, and retain only collision-free pairs
under kinematic replay in MuJoCo. These validated variants expand the
$100~\mathrm{h}$ corpus to approximately $1{,}000~\mathrm{h}$ of
motion--scene training pairs; up to 30 background layouts per scene
and online left--right mirroring ($p=0.5$) further diversify the
geometry during training.

\subsection{Modular Planner--Tracker Architecture}
\label{sec:system}

\subsubsection{Perception-Conditioned Flow Planner}
\label{sec:planner_pretraining}

\textbf{Motion and conditioning.}
We use the following $50~\mathrm{Hz}$ representation for motion frame $i$:
\begin{equation}
\mathbf{s}_i =
\left[
h_i,
(\mathbf{g}^{b}_i)^\top,
(\mathbf{v}^{n}_{i,xy})^\top,
\omega^{n}_{i,z},
\mathbf{q}_i^\top,
\dot{\mathbf{q}}_i^\top
\right]^\top
\in\mathbb{R}^{65},
\label{eq:planner_motion_state}
\end{equation}
where $h_i$ is the root height; $\mathbf{g}^{b}_i$ is gravity expressed in the root frame and encodes root tilt but not yaw; $\mathbf{v}^{n}_{i,xy}$ and $\omega^{n}_{i,z}$ are the planar root velocity and yaw rate in a yaw-aligned navigation frame; and $\mathbf{q}_i,\dot{\mathbf{q}}_i$ are the positions and velocities of the 29 actuated joints. Given four history frames $\mathbf{H}_i=[\mathbf{s}_{i-3},\ldots,\mathbf{s}_i]$, the planner predicts $H=25$ future frames, or $0.5~\mathrm{s}$ of motion. It is additionally conditioned on a robot-centric destination $\mathbf{c}_g$ and a torso-centered three-layer elevation map $\mathbf{E}_i\in\mathbb{R}^{3\times31\times61}$ that jointly encodes supporting surfaces, lateral obstructions, and overhead clearance.

\textbf{Conditional flow matching.}
We adapt a transformer backbone~\cite{wang2026motionbricks,xu2025parc} to jointly process terrain, destination, history, noisy-motion, and flow-time tokens. Let $\widetilde{\mathbf{Y}}\in\mathbb{R}^{H\times65}$ be a normalized ground-truth motion chunk, $\boldsymbol{\epsilon}\sim\mathcal{N}(\mathbf{0},\mathbf{I})$, and $t\sim\mathcal{U}(0,1)$. Using the convention that $t=0$ denotes data and $t=1$ denotes noise,
we define
\begin{equation}
\mathbf{X}_t=(1-t)\widetilde{\mathbf{Y}}+t\boldsymbol{\epsilon},
\qquad
\mathbf{u}^{\star}=\boldsymbol{\epsilon}-\widetilde{\mathbf{Y}}.
\label{eq:planner_flow_path}
\end{equation}
Let $\mathcal{C}=(\mathbf{H},\mathbf{E},\mathbf{c}_g)$ denote the
conditioning inputs. The flow-matching objective is
\begin{equation}
\mathcal{L}_{\mathrm{FM}}
=
\mathbb{E}\!\left[
\frac{1}{65H}\sum_{j,k}
\rho\!\left(
v_{\theta,jk}(\mathbf{X}_t,t;\mathcal{C})
-u^\star_{jk}
\right)
\right],
\label{eq:planner_flow_loss}
\end{equation}
where the sum spans $j=1,\ldots,H$ and $k=1,\ldots,65$, and $\rho$
is the smooth-$L_1$ loss. At inference, generation starts from
Gaussian noise and integrates the learned vector field from $t=1$
to $t=0$.

\textbf{Motion and geometry regularization.}
We form the denoised estimate
\begin{equation}
\widehat{\mathbf{Y}}
=\mathbf{X}_t-t\,v_\theta(\mathbf{X}_t,t;\mathbf{H},\mathbf{E},\mathbf{c}_g),
\label{eq:planner_clean_estimate}
\end{equation}
and decode it through differentiable forward kinematics. Auxiliary losses regularize root-state and joint-velocity reconstruction, position--velocity finite-difference consistency, foot sliding, body-keypoint jerk, and oriented-box penetration. We additionally precompute HumanoidPF guidance~\cite{xue2026cat} from the scene signed-distance field and destination, and sample it at 11 anchors on the pelvis, torso, head, shoulders, palms, knees, and feet. The resulting loss $\mathcal{L}_{\mathrm{PF}}$ repels anchors within $0.20~\mathrm{m}$ of obstacles and penalizes anchor motion opposing the local guidance direction. HumanoidPF is used only for training and adds neither planner observations nor runtime computation. The complete pre-training objective is
\begin{align}
\mathcal{L}_{\mathrm{pre}}
={}&
\mathcal{L}_{\mathrm{FM}}
+2\mathcal{L}_{\mathrm{root}}
+2\mathcal{L}_{\mathrm{jvel}}
+0.05\mathcal{L}_{\mathrm{fd}}
+0.02\mathcal{L}_{\mathrm{slide}}
\nonumber\\
&+0.05\mathcal{L}_{\mathrm{jerk}}
+5\mathcal{L}_{\mathrm{box}}
+\mathcal{L}_{\mathrm{PF}}.
\label{eq:planner_total_pretrain_loss}
\end{align}

\subsubsection{Perception-Enhanced General Tracker}
\label{sec:controller_perception_augmentation}

We use the whole-body interface of ScaleBFM~\cite{zeng2026scaling}, pre-trained on a separate 1,000 h general-motion corpus for a 29-DoF G1 carrying the AGX backpack. ScaleBFM uses proprioception--action tokens as cross-attention queries and motion-reference tokens as keys and values. We append eight terrain tokens encoded by a convolutional neural network (CNN) to its motion-reference token sequence, without changing the architecture of the pretrained transformer or action head. At $50~\mathrm{Hz}$, the tracker maps three-frame proprioceptive and action histories and references for 14 body links to 29-dimensional residual joint-position commands.

We adapt the tracker with asymmetric actor--critic proximal policy optimization (PPO), using whole-body tracking objectives together with safety and smoothness regularization. We first optimize the terrain encoder and critic for 200 iterations and then fine-tune the complete policy with dynamics, observation, and external-perturbation randomization. This adaptation uses reference motions from the scene-aligned dataset, not planner-generated rollouts. The resulting general tracking interface therefore accepts the planner's references directly, without behavior-specific experts or joint planner--tracker optimization.

\subsection{Planner-Side Refinement}
\label{sec:post_training}
\subsubsection{Real-Time Chunking}
\label{sec:planner_rtc_finetuning}

Independent receding-horizon samples may become discontinuous at chunk
boundaries. We therefore adopt the action-prior denoising of Soft
RTC~\cite{liu2026action}, extending hard training-time
conditioning~\cite{black2025training}. We sample
$d\in\{0,\ldots,4\}$ with $p(d=k)\propto e^{-k}$ and set
$e_d=\min\{H,5,\lceil2d\rceil\}$. For token $j$, we define
\begin{equation}
\begin{aligned}
w_j(d)
&=\operatorname{clip}_{[0,1]}
  \left(\frac{e_d-j+1}{e_d-d+1}\right),\\
a_j&=1-w_j(d), \qquad \tau_j=a_jt,\\
\mathbf{X}_{\tau,j}
&=(1-\tau_j)\widetilde{\mathbf{Y}}_j
  +\tau_j\boldsymbol{\epsilon}_j .
\end{aligned}
\label{eq:rtc_corruption}
\end{equation}
Here, $\operatorname{clip}_{[0,1]}$ implements the endpoint-excluding
linear taper. Ground-truth states provide the training prior, leaving
committed tokens clean, transition tokens partially corrupted, and
future tokens under standard flow matching. At inference, the aligned
suffix of the previous chunk provides the prior under the same
token-wise blending rule. We weight token $j$ in
Eq.~\eqref{eq:planner_flow_loss} by $a_j$ and normalize by
$65\sum_j a_j$. Auxiliary reconstruction uses
$\widehat{\mathbf{Y}}^{\mathrm{RTC}}_j
=\mathbf{X}_{\tau,j}-\tau_j\mathbf{v}_{\theta,j}$ with the same terms
as Eq.~\eqref{eq:planner_total_pretrain_loss}. Setting $d=0$ recovers
the standard objective.

\subsubsection{Closed-Loop RL with a Frozen Tracker}
\label{sec:planner_rl_finetuning}

We execute RTC-generated chunks through the frozen perceptive tracker and
update only the planner. Following ReinFlow~\cite{zhang2025reinflow}, a learned
noise scale makes the discretized reverse-flow transitions Gaussian and hence
tractable for PPO. Transitions within a chunk share its planner-level
advantage, and RTC remains active during rollouts.

At the $50~\mathrm{Hz}$ tracker rate, one reward is shared across all scenes
and behaviors:
\begin{equation}
\begin{aligned}
r_t
&=\Delta t\left(
r_t^{\mathrm{goal}}+r_t^{\mathrm{loco}}-c_t^{\mathrm{obs}}
\right),
\qquad \Delta t=0.02~\mathrm{s},\\
r_t^{\mathrm{goal}}
&=0.75r_t^{\mathrm{head}}
+2r_t^{\mathrm{prog}}
+4r_t^{\mathrm{arr}}\\
&\quad
+0.5r_t^{\mathrm{speed}}
+r_t^{\mathrm{hold}}
+0.1r_t^{\mathrm{alive}},\\
r_t^{\mathrm{loco}}
&=0.4r_t^{\mathrm{gait}}
-0.005c_t^{\mathrm{act}}
-0.2c_t^{\mathrm{slip}}
-2c_t^{\mathrm{flight}},\\
c_t^{\mathrm{obs}}
&=4c_t^{\mathrm{contact}}+c_t^{\mathrm{dist}}.
\end{aligned}
\label{eq:planner_rl_reward}
\end{equation}
These terms encode goal reaching and stable stopping, feasible support
transitions, and collision avoidance; $c_t^{\mathrm{dist}}$ provides dense
pre-contact supervision from the clearances between selected hand and foot
points and the environment. During RL,
arrival requires remaining within $0.35~\mathrm{m}$ of the destination below
$0.3~\mathrm{m/s}$ for $0.2~\mathrm{s}$.

Each chunk controls up to $C=8$ tracker steps, so rewards and advantages are
computed at planner boundaries. For decision $j$ spanning $n_j\leq C$ steps,
$\bar r_j=\sum_{i=0}^{n_j-1}\gamma_c^i r_{t_j+i}$ and
\begin{equation}
\begin{aligned}
\delta_j
&=\bar r_j+\gamma_c^{n_j}(1-m_j^{\mathrm{term}})V_{j+1}-V_j,\\
A_j
&=\delta_j+\gamma_c^{n_j}\lambda_{\mathrm{GAE}}
(1-m_j^{\mathrm{term}})(1-m_j^{\mathrm{trunc}})A_{j+1}.
\end{aligned}
\label{eq:planner_semimdp_gae}
\end{equation}
Here $m_j^{\mathrm{term}}$ and $m_j^{\mathrm{trunc}}$ denote true termination
and time-limit or rollout truncation, respectively. We use $\gamma_c=0.98$ and
$\lambda_{\mathrm{GAE}}=0.95$; true terminals suppress bootstrapping, whereas
truncations retain the next-state bootstrap but stop advantage recursion.
Advantages are batch-normalized.

We train for 1{,}000 iterations in 128 parallel environments spanning 16
scenes and 1{,}720 clips. Freezing the tracker exposes the planner to
execution-induced deviations without changing the planner--tracker interface.

\subsection{Egocentric Perception and Onboard Deployment}
\label{sec:deploy}

\begin{figure}[t]
\centering
\includegraphics[width=\columnwidth]{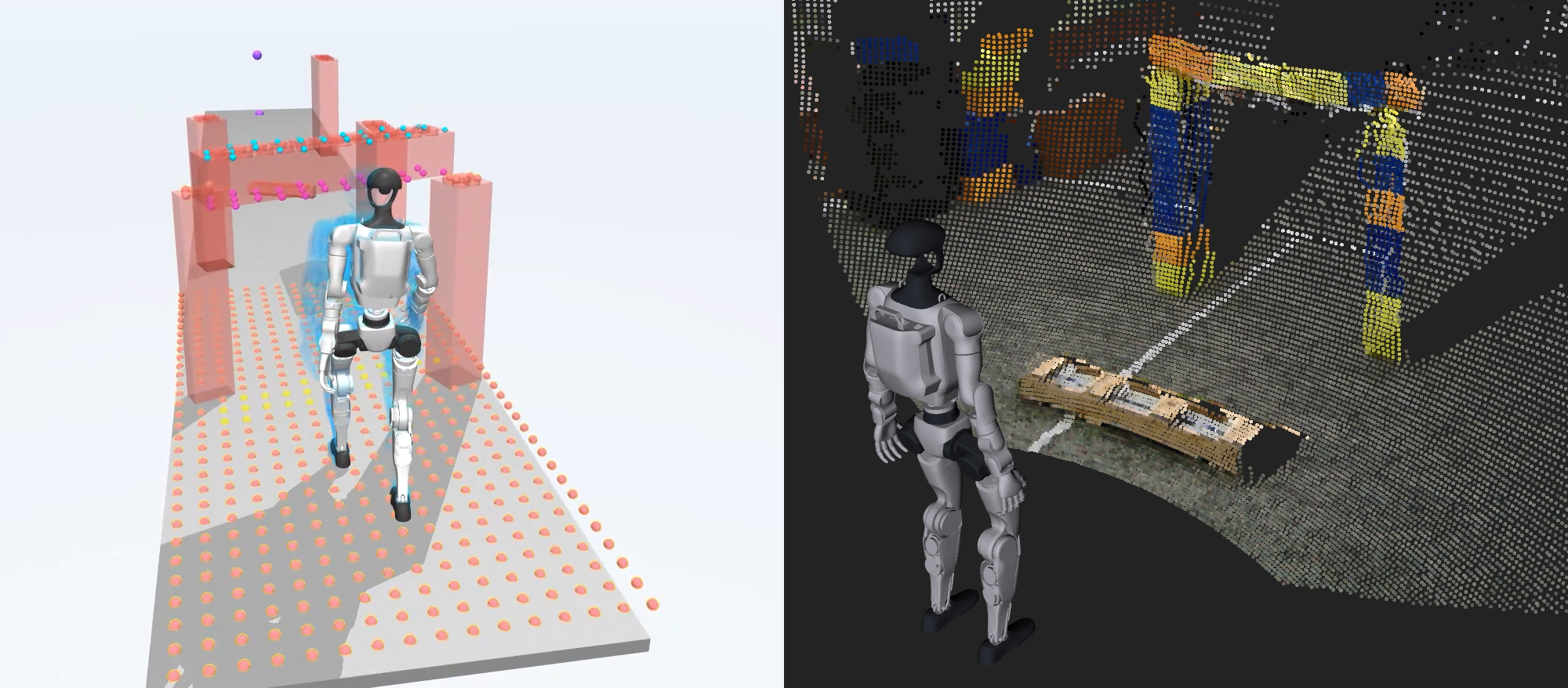}
\caption{The shared three-layer geometric representation in simulation (left) and reconstructed online from onboard LiDAR without a prebuilt map (right).}
\label{fig:sim2real}
\end{figure}

The complete stack runs onboard a Unitree G1 equipped with a Jetson AGX Orin and a Manifold Tech Odin module, without a prebuilt map or offboard computation. Registered point clouds and odometry are fused online into a robocentric 3D occupancy grid maintained by ROG-Map~\cite{ren2023rogmap}. We extract the same torso-centered three-layer elevation map used in training: each vertical voxel column yields the highest occupied surface, an obstacle underside supported by observed free space beneath it, and the supporting surface below that clearance. Together, the three layers encode supporting geometry, lateral blockage, and overhead clearance; unknown voxels are never treated as free. Fig.~\ref{fig:sim2real} contrasts this representation in simulation and onboard operation.

The extracted surfaces are cached and resampled in the torso frame at $20~\mathrm{Hz}$ using the latest odometry, reducing pose staleness between point-cloud updates. The planner runs in a dedicated process with TensorRT FP16 at $6.25~\mathrm{Hz}$, the tracker runs on the onboard CPU at $50~\mathrm{Hz}$, and joint targets are published to the robot at $500~\mathrm{Hz}$.

\begin{table*}[t]
\caption{Quantitative comparison, data scaling, and ablations on
held-out cluttered scenes. Scaling rows report mean $\pm$ standard
deviation over three independent training seeds; all other rows report
single checkpoints.}
\label{tab:overall_results}
\centering
\begin{tabular}{@{}lccccc@{}}
\toprule
Method
& Succ.$\uparrow$
& CF-Succ.$\uparrow$
& Fall$\downarrow$
& Contact/Path$\downarrow$
& FSlip$\downarrow$ \\
\midrule
CAT (released, zero-shot)~\cite{xue2026cat}
& 0.7027
& 0.1400
& 0.2013
& 0.3192
& 0.1738 \\

Zhang et al.-style pipeline (our impl., $12~\mathrm{h}$)%
~\cite{zhang2026learningwholebodyhumanoidlocomotion}
& 0.8600
& 0.1920
& 0.0080
& 0.5680
& 0.0120 \\
\midrule

Ours w/o RTC
& 0.9267
& 0.2453
& 0.0307
& 0.4038
& 0.0119 \\

Ours w/o RL post-training
& 0.9440
& 0.2640
& 0.0293
& 0.1985
& 0.0185 \\

Ours w/o tracker perception
& 0.8440
& 0.1840
& 0.0693
& 0.2520
& 0.0147 \\

Ours w/o PF loss
& 0.8480
& 0.2013
& 0.0253
& 0.1538
& 0.0134 \\
\midrule

Ours, $6~\mathrm{h}$
& $0.843\pm0.004$
& $0.481\pm0.007$
& $0.008\pm0.012$
& $0.357\pm0.051$
& $0.0124\pm0.0005$ \\

Ours, $12~\mathrm{h}$
& $0.900\pm0.005$
& $0.486\pm0.003$
& $0.016\pm0.001$
& $0.123\pm0.018$
& $0.0125\pm0.0006$ \\

Ours, $24~\mathrm{h}$
& $0.923\pm0.003$
& $0.578\pm0.005$
& $0.011\pm0.003$
& $0.080\pm0.003$
& $0.0112\pm0.0010$ \\

Ours, $48~\mathrm{h}$
& $0.934\pm0.007$
& $0.643\pm0.004$
& $0.010\pm0.005$
& $0.060\pm0.001$
& $0.0115\pm0.0006$ \\

Ours, $100~\mathrm{h}$ (w/o aug.)
& $0.964\pm0.007$
& $0.689\pm0.004$
& $\mathbf{0.006}\pm0.002$
& $0.050\pm0.001$
& $\mathbf{0.0108}\pm0.0021$ \\
\midrule

Ours, $100~\mathrm{h}$ + aug. (final)
& \textbf{0.9867}
& \textbf{0.7027}
& 0.0093
& \textbf{0.0422}
& 0.0113 \\
\bottomrule
\end{tabular}
\end{table*}


\section{Experiments}

Our experiments evaluate two central claims: (i) PASSAGE can autonomously
select, compose, and execute whole-body traversal behaviors across
ground-level, lateral, and overhead constraints using a single
destination-conditioned planner and a perceptive general tracker, without
skill labels or behavior-specific controllers; and (ii) increasing
scene-aligned training data yields a robust positive empirical trend in
closed-loop generalization to held-out scenes. We further isolate the effects
of RTC and closed-loop planner post-training and quantify the complete system
under fully onboard real-world execution.

\subsection{Evaluation Protocol}

All simulation experiments run in MuJoCo~\cite{todorov2012mujoco} on held-out
instances from the procedural generator in Sec.~\ref{sec:data}. The test set
contains 150 scenes, with 50 at each of three difficulty levels. For every
method or configuration, we conduct five rollouts per scene using identical
scenes, start--destination pairs, and rollout seeds, yielding 750 episodes.
Each episode runs at $50~\mathrm{Hz}$ for at most $60~\mathrm{s}$. Success
(Succ.) requires reaching within $0.5~\mathrm{m}$ of the destination before a
fall or timeout; contact-free success (CF-Succ.) additionally requires zero
robot--obstacle contact. Fall is the fraction of episodes terminated by low
root height or excessive body tilt. Contact/Path is obstacle-contact duration
divided by planar distance traveled, in $\mathrm{s/m}$. Foot Slip is the
temporal mean of
$\sum_{i\in\mathcal{C}_t}\lVert\mathbf{v}^{xy}_{i,t}\rVert_2^2$,
in $\mathrm{m^2/s^2}$. Metrics are first averaged within each scene and then
macro-averaged across scenes.

\subsection{Overall Closed-Loop Performance}

We compare PASSAGE with two complementary baselines under the common
evaluation protocol. The released CAT generalist~\cite{xue2026cat} is
evaluated without retraining and retains its native interfaces. Although
the test scenes are unseen by both systems, they follow PASSAGE's training
distribution and are encountered by CAT zero-shot; this is therefore a
system-level transfer comparison.

We also implement a Zhang et al.-style diffusion--tracking
pipeline~\cite{zhang2026learningwholebodyhumanoidlocomotion}. Its planner
uses the same unaugmented, nested $12~\mathrm{h}$ subset and three-layer
geometric representation as the corresponding PASSAGE planner. Following
the reported design, it predicts 25 frames from two history frames with two
denoising steps. A task-specific perceptive tracker is trained on the same
subset and subsequently fine-tuned in closed loop with the planner frozen.
We follow reported settings where available and document our choices for
unspecified details. Simulator scene-loading constraints preclude training
this tracker as a general tracker.

Final PASSAGE achieves $98.7\%$ Succ. and $70.3\%$ CF-Succ., compared
with $70.3\%$ and $14.0\%$ for CAT. The Zhang et al.-style pipeline
achieves $86.0\%$ Succ. and $19.2\%$ CF-Succ., whereas PASSAGE at the
same $12~\mathrm{h}$ planner-data scale averages $90.0\%$ and $48.6\%$
over three training seeds, respectively, and reduces Contact/Path from
$0.5680$ to $0.1230~\mathrm{s/m}$. Thus, both pipelines frequently reach
the destination, but PASSAGE completes substantially more trials without
contact. Because their tracker training and closed-loop refinement also
differ, this comparison evaluates complete pipelines rather than isolating
flow matching from diffusion.

\begin{figure}[t]
    \centering
    \includegraphics[width=\columnwidth]{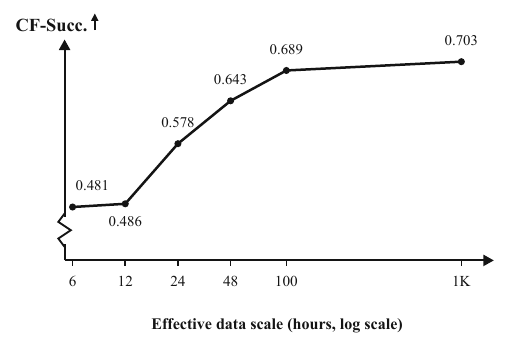}
    \caption{Closed-loop CF-Succ. across scene-aligned training-data scales.
    Points from 6 to $100~\mathrm{h}$ are means over three independent
    training seeds using nested unaugmented subsets; the
    $1{,}000~\mathrm{h}$ point is the single final checkpoint obtained through
    $1{:}10$ scene augmentation of the $100~\mathrm{h}$ corpus.}
    \label{fig:scaling_cf_succ}
\end{figure}

\subsection{Scaling Scene-Aligned Training Data}

We examine captured-data scaling separately from scene augmentation. At each
of 6, 12, 24, 48, and $100~\mathrm{h}$, we train three planners with
independent seeds and report mean $\pm$ standard deviation. All runs use the
same architecture, optimization budget, RTC and PF objectives, closed-loop
RL post-training, frozen tracker, and evaluation episodes; across scales,
only the planner pre-training subset changes. The final augmented model
additionally applies $1{:}10$ scene augmentation to the full
$100~\mathrm{h}$ corpus, producing approximately $1{,}000~\mathrm{h}$ of
effective motion--scene pair duration without additional human capture.

Across three seeds, mean Succ. increases from $(84.3\pm0.4)\%$ at
$6~\mathrm{h}$ to $(96.4\pm0.7)\%$ at $100~\mathrm{h}$, while mean
CF-Succ. rises from $(48.1\pm0.7)\%$ to $(68.9\pm0.4)\%$.
Contact/Path decreases from $0.3565\pm0.0512$ to
$0.0499\pm0.0013~\mathrm{s/m}$. The $6$--$12~\mathrm{h}$ CF-Succ.
change lies within seed variation, whereas Succ. and Contact/Path improve
monotonically for every seed. Fall and Foot Slip show no consistent
monotonic trend; the strongest gains therefore occur in goal reaching and
collision avoidance.

Relative to the three-seed mean of the $100~\mathrm{h}$ no-augmentation
control, the single augmented model reaches $98.7\%$ Succ. and $70.3\%$
CF-Succ., with $0.0422~\mathrm{s/m}$ Contact/Path. These are observed
differences of $2.2$ and $1.3$ percentage points in Succ. and CF-Succ.,
respectively, and a $15.4\%$ reduction in contact exposure. Thus, the
6--$100~\mathrm{h}$ study supports a robust positive empirical trend with
captured-data scale. We analyze the smaller augmentation gain separately
because it reuses the same $100~\mathrm{h}$ of captured motion
(Fig.~\ref{fig:scaling_cf_succ}). The subsets are nested at the trajectory
level, and all test-scene seeds are disjoint from those used for
pre-training, scene augmentation, RL post-training, and model selection.

\begin{figure*}[t]
\centering
\begin{tikzpicture}[panel/.style={anchor=north west, xshift=3pt, yshift=-3pt,
        font=\bfseries\small, text=white, fill=black, fill opacity=0.55,
        text opacity=1, inner sep=2.5pt, rounded corners=1.5pt}]
    \node[anchor=south west, inner sep=0] (img)
        {\includegraphics[width=\textwidth]{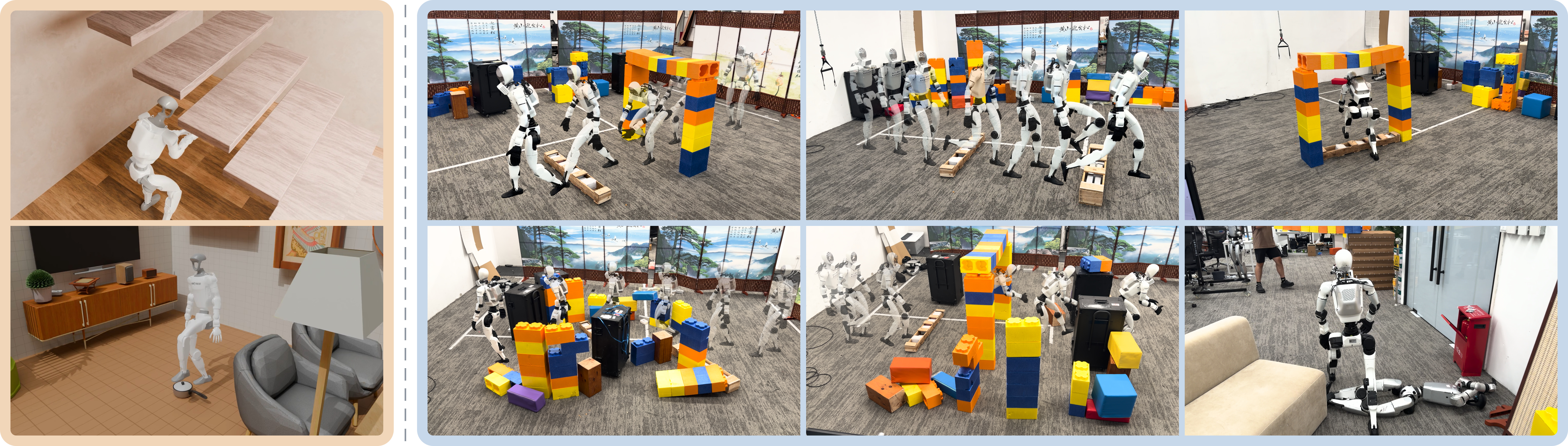}};
    \begin{scope}[x={(img.south east)}, y={(img.north west)}]
        \node[panel] at (0.007, 0.977) {(a)};
        \node[panel] at (0.007, 0.494) {(b)};
        \node[panel] at (0.273, 0.977) {(c)};
        \node[panel] at (0.514, 0.977) {(d)};
        \node[panel] at (0.756, 0.977) {(e)};
        \node[panel] at (0.273, 0.494) {(f)};
        \node[panel] at (0.514, 0.494) {(g)};
        \node[panel] at (0.756, 0.494) {(h)};
    \end{scope}
\end{tikzpicture}
\caption{Qualitative simulated \textbf{(a)--(b)} and real-world
\textbf{(c)--(h)} rollouts: \textbf{(a)}~duck-under; \textbf{(b)}~step-over;
\textbf{(c)}~step-over followed by duck-under; \textbf{(d)}~repeated
step-over; \textbf{(e)}~concurrent step-over and duck-under;
\textbf{(f)}~squeeze-through; \textbf{(g)}~dense-clutter weaving; and
\textbf{(h)}~stepping over a fallen robot. Real-world trials use egocentric
perception and fully onboard computation; overlays show successive poses.}
\label{fig:realworld}
\end{figure*}

\subsection{Component Ablations}

Table~\ref{tab:overall_results} isolates the contributions of RTC,
closed-loop reinforcement learning (RL) post-training, tracker-side
perception augmentation, and potential-field (PF) regularization. All
variants retain the full planner training dataset and change only the
indicated component relative to the final system. For planner-side
ablations, the component is removed from every stage in which it is used.
For the tracker ablation, we replace the perception-enhanced tracker with
the base ScaleBFM checkpoint while leaving the planner and
planner--tracker interface unchanged.

\textbf{Real-time chunking.}
Removing RTC reduces Succ. from $98.7\%$ to $92.7\%$ and CF-Succ. from
$70.3\%$ to $24.5\%$, while increasing Contact/Path nearly tenfold from
$0.0422$ to $0.4038~\mathrm{s/m}$. This safety degradation is consistent
with RTC reducing geometry-relative discontinuities across replans, which
can otherwise turn small reference shifts into obstacle contact.

\textbf{Closed-loop planner post-training.}
Removing RL post-training reduces Succ. to $94.4\%$ and CF-Succ. to
$26.4\%$, while increasing Contact/Path to $0.1985~\mathrm{s/m}$ and
Foot Slip to $0.0185$. These changes indicate that optimizing the planner
through rollouts under the frozen tracker improves realized goal reaching,
collision avoidance, and execution quality. The improvement is consistent
with reducing the mismatch between offline planner outputs and
tracker-realized states without altering the modular planner--tracker
interface.

\textbf{Tracker-side perception augmentation.}
Replacing the perception-enhanced tracker with the base ScaleBFM tracker
reduces Succ. from $98.7\%$ to $84.4\%$ and CF-Succ. from $70.3\%$ to
$18.4\%$, while increasing Fall from $0.9\%$ to $6.9\%$. Because the
planner and planner--tracker interface remain unchanged, this comparison
measures the net contribution of tracker-side perception augmentation.
The degradation is consistent with local geometric feedback helping
correct tracking errors under tight clearances.

\textbf{Potential-field regularization.}
Removing the PF loss reduces Succ. to $84.8\%$ and CF-Succ. to $20.1\%$,
while increasing Contact/Path from $0.0422$ to $0.1538~\mathrm{s/m}$.
This safety degradation is consistent with PF acting as a geometric prior
that discourages residual scene--motion interpenetration introduced by
motion capture, retargeting, and scene augmentation.

Together with the preceding scaling study, these results support a
complementary division of roles: scene-aligned demonstrations provide the
geometry-conditioned behavioral repertoire; PF supplies an explicit
collision-avoidance prior; RTC promotes consistency across successive
plans; closed-loop RL adapts the planner to tracker-realized motion; and
tracker-side perception augmentation equips the execution layer with local
geometric correction. All component ablations are conducted in simulation,
while the following experiments evaluate the resulting complete stack on
hardware.

\subsection{Onboard Real-World Evaluation}

We evaluate the same planner--tracker pair across 50 physical layouts using
destination-only commands, without behavior labels or an explicit skill
selector. Table~\ref{tab:real_robot} summarizes the results.

\begin{table}[t]
    \centering
    \caption{Real-world traversal with the complete onboard system. Each
    scenario contains ten distinct physical layouts, each attempted once;
    all attempts are included in $N$. Succ. denotes reaching the designated
    finish region beyond the obstacle layout without a fall or safety
    intervention. CF-Succ. additionally requires no robot--obstacle contact,
    as manually annotated by human observers.}
    \label{tab:real_robot}
    \begin{tabular}{@{}lccc@{}}
        \toprule
        Scenario & $N$ & Succ. & CF-Succ. \\
        \midrule
        Step-over              & 10 & 10/10 & 10/10 \\
        Duck-under             & 10 & 10/10 &  9/10 \\
        Squeeze-through        & 10 & 10/10 &  9/10 \\
        Concurrent over--under & 10 & 10/10 &  9/10 \\
        Sequential mixed       & 10 & 10/10 &  8/10 \\
        \midrule
        \textbf{Overall} & \textbf{50} & \textbf{50/50}
        & \textbf{45/50} \\
        \bottomrule
    \end{tabular}
\end{table}

PASSAGE reaches the finish region without a fall or safety intervention in
all 50 trials, with 45/50 ($90\%$) contact-free executions. It completes all
20 concurrent and sequential layouts, of which 17 are contact-free,
including 8/10 sequential mixed courses. Using a distinct physical layout
in every trial extends the evaluation across 50 geometric configurations.
Together with the geometry-conditioned behaviors in
Fig.~\ref{fig:realworld}, these results provide task-level evidence that a
single onboard system can select and compose whole-body traversal behaviors
without an explicit skill selector. The five contact-bearing completions
indicate that collision avoidance, rather than course completion, remains
the principal limitation in the tested layouts.

\section{Conclusion}

We presented PASSAGE, a scalable, perception-conditioned framework for
destination-directed humanoid traversal in cluttered environments. Our VR and
inertial motion-capture pipeline enabled the collection of $100~\mathrm{h}$ of
scene-aligned human demonstrations, expanded through validated scene
augmentation to approximately $1{,}000~\mathrm{h}$ of motion--scene training
pairs. From these data, a single flow-matching planner learns to select and
compose whole-body behaviors across ground-level, lateral, and overhead
constraints, while a perceptive general tracker executes them without skill
labels or behavior-specific controllers. RTC promotes consistency across
replanning steps, and planner-side RL post-training with a frozen tracker
improves closed-loop performance. Across three independent training seeds,
scaling captured data from 6 to $100~\mathrm{h}$ increases mean goal-reaching
success from $84.3\%$ to $96.4\%$ and contact-free success from $48.1\%$ to
$68.9\%$, while reducing Contact/Path from $0.3565$ to
$0.0499~\mathrm{s/m}$. With egocentric perception and fully onboard
computation, PASSAGE reaches the destination in all 50 real-world trials, of
which 45 are contact-free.

Future work will broaden both the behavioral and perceptual scope of humanoid
traversal. More diverse demonstrations could extend the learned repertoire to
ascending and descending stairs, climbing onto platforms, and vaulting over
obstacles. Complementing geometric sensing with RGB observations may improve
perception of transparent surfaces, thin cables, and other structures that
remain challenging for LiDAR alone. True to its name, PASSAGE represents one
step toward humanoids that traverse human environments with increasingly
broad, human-like adaptability.


\bibliographystyle{IEEEtran}
\bibliography{IEEEabrv,library}

\appendices

\section{Planner Architecture, Training, and Refinement Details}
\label{app:planner}

\subsection{Architecture and Conditioning}
\label{app:planner_architecture}

The 65-D state follows Eq.~\eqref{eq:planner_motion_state}. Root velocities are
expressed in the yaw-aligned navigation frame obtained from the pelvis frame by
removing roll and pitch at each sample; absolute horizontal position and yaw
are omitted. The robot-centric destination $\mathbf{c}_g$ is expressed in the
current torso-yaw frame and embedded by an MLP. History and future motion use
separate channel-wise statistics computed from the training split, with
standard deviations lower-bounded by $10^{-3}$. The planner receives the same
torso-centered, yaw-aligned three-layer map
$\mathbf{E}\in\mathbb{R}^{3\times31\times61}$ used by the tracker. It follows
the torso-relative downward-depth convention in
Appendix~\ref{app:deployment} and is standardized with training-set statistics
before tokenization.
Table~\ref{tab:app_planner_architecture} summarizes the planner architecture.

We adopt the 16-block, width-1024 scale of the MotionBricks pose
backbone~\cite{wang2026motionbricks}, but use eight attention heads and
directly predict a continuous $25\times65$ vector field. We do not use
MotionBricks' discrete
codebook or root--pose latent decomposition. The terrain, destination,
history, and noisy-motion tokenization is inspired by PARC~\cite{xu2025parc},
and RTC follows Soft RTC~\cite{liu2026action}.

\subsection{Offline Training and Flow Inference}
\label{app:planner_pretraining}

We sample $u\sim\mathcal{U}(0,1)$, set
$t=\operatorname{clip}(u,10^{-5},1-10^{-5})$, and draw
$\boldsymbol{\epsilon}\sim\mathcal{N}(\mathbf{0},\mathbf{I})$. Gaussian noise
with standard deviation 0.01 is added to normalized history and terrain
inputs; the complete history is masked with probability 0.15, and left--right
mirroring is applied with probability 0.5. All Smooth-$L_1$ terms use
transition parameter $\beta=1$. After standard offline pre-training, we
fine-tune the same planner with the main-text RTC objective. During this RTC
fine-tuning stage, fully committed tokens are excluded from reconstruction and
kinematic losses, whereas the potential-field loss is evaluated on all
predicted frames.

The root and joint-velocity losses act on the first seven and final 29 state
channels, respectively. After unnormalization, the finite-difference loss
matches predicted joint velocity to
$(\mathbf{q}_i-\mathbf{q}_{i-1})/\Delta t$, where
$\Delta t=0.02~\mathrm{s}$. Foot sliding is penalized when
consecutive ankle positions lie within a soft $0.08~\mathrm{m}$ support band.
Body-keypoint jerk is evaluated at the pelvis, torso, ankles, and wrists with a
$750~\mathrm{m/s^3}$ hinge threshold, and the oriented-box loss applies a
$0.20~\mathrm{m}$ margin to the same six points.

For potential-field regularization, the 11 anchors are the pelvis, torso,
head, bilateral knees, feet, shoulders, and palms, implemented at the
ankle-roll and wrist-yaw links. Let $d_{\mathrm{sdf}}$ be signed distance,
$\widehat{\mathbf{G}}$ the normalized guidance direction,
$\Delta\mathbf{p}$ an anchor displacement, and
$r_{\mathrm{PF}}=0.20~\mathrm{m}$. We use
\begin{equation}
\begin{aligned}
\mathcal{L}_{\mathrm{rep}}
&=\left\langle
\frac{(r_{\mathrm{PF}}-d_{\mathrm{sdf}})^2}{2r_{\mathrm{PF}}}
\right\rangle_{d_{\mathrm{sdf}}<r_{\mathrm{PF}}},\\
\mathcal{L}_{\mathrm{dir}}
&=\left\langle
[-\widehat{\Delta\mathbf{p}}^\top\widehat{\mathbf{G}}]_+
\operatorname{clip}\!\left(
\frac{\|\Delta\mathbf{p}\|}{0.05~\mathrm{m}},0,1\right)
\right\rangle,
\end{aligned}
\label{eq:app_pf_loss}
\end{equation}
where $[x]_+=\max(x,0)$ and $\widehat{\Delta\mathbf{p}}=
\Delta\mathbf{p}/(\|\Delta\mathbf{p}\|_2+10^{-8}~\mathrm{m})$. The first
average is over anchors with $d_{\mathrm{sdf}}<r_{\mathrm{PF}}$ and is set to
zero when none are active; the second is over
consecutive anchor displacements. We set
$\mathcal{L}_{\mathrm{PF}}=\mathcal{L}_{\mathrm{rep}}+
\mathcal{L}_{\mathrm{dir}}$. Scene fields are trilinearly interpolated and
used only during training.

At inference, generation starts from Gaussian noise and numerically integrates
the learned vector field from $t=1$ to $t=0$. The planner outputs 25 frames;
the first eight are executed before replanning, consistent with
$6.25~\mathrm{Hz}$ planning and $50~\mathrm{Hz}$ tracking.

\subsection{Real-Time Chunking and Tracker Interface}
\label{app:planner_rtc_interface}

At inference, the temporally aligned unexecuted suffix of the preceding chunk
supplies the RTC prior under the token-wise blending rule defined in the main
text. If no valid suffix exists, including the first plan after reset, we set
$d=0$. The four-frame history is subsequently updated from tracker-executed
rather than planned states.

After denormalization, each generated chunk is re-anchored to the latest
measured pelvis pose. For successive generated frames and
$\Delta t=0.02~\mathrm{s}$, the planar root trajectory is reconstructed as
\begin{equation}
\begin{aligned}
\mathbf{p}_{j+1,xy}^{w}
&=\mathbf{p}_{j,xy}^{w}
+\Delta t\,R_{xy}(\psi_j)\mathbf{v}_{j,xy}^{n},\\
\psi_{j+1}&=\psi_j+\Delta t\,\omega_{j,z}^{n},
\end{aligned}
\label{eq:app_root_decode}
\end{equation}
where $R_{xy}(\psi)$ denotes planar yaw rotation. Predicted root height and
gravity recover vertical position and tilt, and forward kinematics produces the
14-link references defined in Appendix~\ref{app:tracker}.

\begin{table}[t]
\caption{Flow-planner architecture and tokenization.}
\label{tab:app_planner_architecture}
\centering
\small
\setlength{\tabcolsep}{4pt}
\begin{tabularx}{\columnwidth}{@{}lX@{}}
\toprule
Component & Configuration \\
\midrule
Time / goal & Sinusoidal flow-time embeddings and an MLP goal embedding,
projected to 1024-D \\
Terrain & Three stride-1 $3\times3$ Conv--GN--SiLU layers
($3\!\rightarrow\!32\!\rightarrow\!64\!\rightarrow\!64$; GN groups
$4/8/8$), replicate padding to $33\times63$, and a $3\times3$, stride-3
projection to 1024-D; 231 tokens \\
Motion & Separate shared two-layer MLPs for four history and 25 noisy-future
tokens \\
Sequence & Terrain, history, and future tokens use separate fixed positional
encodings and learned modality embeddings \\
Transformer & 16 pre-norm blocks, width 1024, eight heads, FFN width 4096,
GELU, dropout 0.1, and global bidirectional attention \\
Output & LayerNorm followed by a
$1024\!\rightarrow\!1024\!\rightarrow\!65$ head on the final 25 tokens \\
\bottomrule
\end{tabularx}
\end{table}

\subsection{Planner-Side RL Post-Training}
\label{app:planner_rl}

We execute RTC-generated chunks through the frozen perceptive tracker and
update only the planner. Following ReinFlow~\cite{zhang2025reinflow}, a learned
noise scale makes the discretized reverse-flow transitions Gaussian and
tractable for PPO. Transitions belonging to one generated chunk share the
planner-boundary advantage defined in the main text; the tracker remains frozen
throughout.

Let $\Delta\mathbf{g}_t$ be the horizontal pelvis-to-goal displacement,
$d_t=\|\Delta\mathbf{g}_t\|_2$, and
$\widehat{\mathbf{g}}_t=\Delta\mathbf{g}_t/(d_t+10^{-4}~\mathrm{m})$; we set
$\widehat{\mathbf{g}}_t=\mathbf{0}$ at the goal. Let $\mathbf{v}_t$ be
horizontal pelvis velocity and $I$ a binary indicator.
Table~\ref{tab:app_rl_atoms} defines the reward atoms whose weights are given
in the main text. Physical quantities are evaluated numerically in SI units;
the scalar reward coefficients absorb the corresponding units.

\begin{table*}[t]
\caption{Planner-side RL reward atoms before the main-text weights and common
$\Delta t$ factor.}
\label{tab:app_rl_atoms}
\centering
\scriptsize
\setlength{\tabcolsep}{4pt}
\begin{tabularx}{\textwidth}{@{}lX@{}}
\toprule
Term & Definition \\
\midrule
$r^{\mathrm{head}}$ &
$\widehat{\mathbf{g}}_t^\top\mathbf{v}_t/
(\|\mathbf{v}_t\|+10^{-4}~\mathrm{m/s})$;
zero at rest \\
$r^{\mathrm{prog}}$ &
$\operatorname{clip}((d_{t-1}-d_t)/\Delta t,
-3~\mathrm{m/s},3~\mathrm{m/s})$ \\
$r^{\mathrm{arr}}$ &
$\Delta t^{-1}$ once the stable-arrival criterion defined in the main text is
first satisfied \\
$r^{\mathrm{speed}}$ &
$\exp[-(\mathbf{v}_t^\top\widehat{\mathbf{g}}_t-v_t^\star)^2/
(0.2~\mathrm{m/s})^2]$, where
$v_t^\star=\min(0.6~\mathrm{m/s},d_t/(1~\mathrm{s}))$ \\
$r^{\mathrm{hold}}$ &
$\exp[-d_t^2/(0.35~\mathrm{m})^2]
\exp[-\|\mathbf{v}_t\|^2/(0.3~\mathrm{m/s})^2]$ \\
$r^{\mathrm{alive}}$ & Nonterminal indicator \\
$r^{\mathrm{gait}}$ &
$0.6I^{\mathrm{single}}+0.2I^{\mathrm{double}}+
0.5I^{\mathrm{single}}I[d_t>0.1~\mathrm{m}]s^{\mathrm{swing}}+
0.5I^{\mathrm{single\mbox{-}landing}}b$, where $s^{\mathrm{swing}}$ is
goal-directed swing-foot speed divided by $0.5~\mathrm{m/s}$ and clipped to
$[0,1]$; $b=0.5/1/0$ for the first, alternating, or repeated-same-foot landing \\
$c^{\mathrm{act}}$ & $\|\mathbf{a}_t-\mathbf{a}_{t-1}\|_2^2$ \\
$c^{\mathrm{slip}}$ &
$\sum_f I^\mathrm{contact}_{f,t}\|\mathbf{v}^{xy}_{f,t}\|_2^2$ \\
$c^{\mathrm{flight}}$ & Indicator that neither foot is in contact \\
$c^{\mathrm{contact}}$ & Number of obstacle-contact channels whose maximum
force norm within the current control interval exceeds $1~\mathrm{N}$ \\
$c^{\mathrm{dist}}$ & Maximum hand/foot-to-obstacle proximity penalty within
$0.10~\mathrm{m}$ \\
\bottomrule
\end{tabularx}
\end{table*}

For $c^{\mathrm{dist}}$, columns less than $0.06~\mathrm{m}$ above nominal
ground are ignored. We use toe, sole, and heel points on both feet and one
endpoint on each hand. For endpoint $k$ and retained vertical column $j$,
\begin{equation}
\begin{aligned}
c_{kj}&=\left[
\|\mathbf{p}_k-\Pi_j(\mathbf{p}_k)\|_2-r_k-r_{\mathrm{cell}}
\right]_+,\\
c^{\mathrm{dist}}&=\max_{k,j}
\mathbf{1}[c_{kj}<0.10~\mathrm{m}]
\exp[-c_{kj}/(0.02~\mathrm{m})],
\end{aligned}
\label{eq:app_endpoint_clearance}
\end{equation}
where $\Pi_j$ projects onto column $j$,
$r_{\mathrm{cell}}=0.0354~\mathrm{m}$, and $r_k$ is $0.012~\mathrm{m}$ for
foot points and $0.035~\mathrm{m}$ for hand points. Foot longitudinal offsets
are $\{0.125,0.040,-0.045\}~\mathrm{m}$ with vertical offset
$-0.029~\mathrm{m}$; hand offsets are $(0.15,\mp0.02,0)~\mathrm{m}$. All
offsets are expressed in their corresponding link-local frames. Taking the
maximum prevents an unsafe endpoint from being diluted by safe ones.

Planner-side RL post-training runs for 1{,}000 iterations in 128 parallel
environments spanning 16 training scenes and 1{,}720 motion clips. We use
$\gamma_c=0.98$ and $\lambda_{\mathrm{GAE}}=0.95$, and batch-normalize the
advantages, as described in the main text.

\section{Perception-Enhanced ScaleBFM Details}
\label{app:tracker}

We use the medium ScaleBFM-compatible Humanoid Transformer
architecture~\cite{zeng2026scaling}, initialized from our checkpoint pretrained
for 20{,}000 iterations on a separate $1{,}000~\mathrm{h}$ Noitom
general-motion corpus for the backpack-equipped 29-DoF G1 described in the main
text. This corpus is distinct from PASSAGE's scene-aligned planner-training
corpus. Perceptive adaptation uses only our scene-aligned traversal references
and no planner-generated rollouts. The actor has four Transformer blocks,
width 256,
four attention heads, and a 256-dimensional SwiGLU feed-forward layer. PASSAGE
adds independently parameterized terrain encoders to the actor and critic; the
deployed actor contains 3.075~M parameters.

\subsection{Tracker Interface and Terrain Conditioning}
\label{app:tracker_interface}

The policy runs at $50~\mathrm{Hz}$. Let $B$ denote the current pelvis frame.
Each actor proprioceptive frame contains
projected gravity, pelvis angular velocity, joint displacement from the nominal
pose, and joint velocity:
\begin{equation}
\mathbf{s}^{p}_t=
[\mathbf{g}^{B}_t,\boldsymbol{\omega}^{B}_t,
\mathbf{q}_t-\mathbf{q}^{\mathrm{nom}},0.05\dot{\mathbf{q}}_t]
\in\mathbb{R}^{64}.
\end{equation}
Let $z_t^p=f_p(\mathbf{s}_t^p)$ and $z_t^a=f_a(\mathbf{a}_t)$ denote the
256-D outputs of the learned proprioception and action tokenizers. The
three-frame proprioceptive--action history comprises three proprioceptive
frames and the two intervening actions, tokenized as
$[z^p_{t-2},z^a_{t-2},z^p_{t-1},z^a_{t-1},z^p_t,e]$, where $e$ is the learned
action-query token.

Let $(\mathbf{p}_i,\mathbf{R}_i)$ and
$(\mathbf{p}^{\star}_{i,t+\delta},\mathbf{R}^{\star}_{i,t+\delta})$ denote
the current and reference poses of link $i$, respectively, where $\delta$ is
measured in $50~\mathrm{Hz}$ control frames. Define
$\mathcal{R}_6(\mathbf{R})=[(\mathbf{R}\mathbf{e}_x)^\top,
(\mathbf{R}\mathbf{e}_z)^\top]^\top\in\mathbb{R}^{6}$. For each future
offset $\delta$, the actor reference is
\begin{equation}
\mathbf{x}^{\mathrm{ref}}_{t,\delta}=
\left[
\left\{
\begin{array}{c}
\mathbf{R}_B^\top(\mathbf{p}^\star_{i,t+\delta}-\mathbf{p}_B)\\
\mathbf{R}_B^\top(\mathbf{p}^\star_{i,t+\delta}-\mathbf{p}_i)\\
\mathcal{R}_6(\mathbf{R}_B^\top\mathbf{R}^\star_{i,t+\delta})\\
\mathcal{R}_6(\mathbf{R}_i^\top\mathbf{R}^\star_{i,t+\delta})
\end{array}
\right\}_{i\in\mathcal{B}},
\delta,\mathbf{m}
\right],
\label{eq:tracker_reference}
\end{equation}
where $|\mathcal{B}|=14$ and $\mathbf{m}=\mathbf{1}_{14}$ is the whole-body
mask. The four link-wise blocks contain $252$ components; appending $\delta$
and the mask yields a 267-D actor token. The actor receives no reference
velocities. The critic omits the mask and appends link linear- and
angular-velocity errors, giving $253+14(3+3)=337$ components per reference
frame. The links are the pelvis; bilateral hip-roll, knee, and ankle-roll
links; torso; and bilateral shoulder-roll, elbow, and wrist-yaw links.

During perceptive adaptation, actor offsets are
$\{0,1,2,3,4,\delta_{\mathrm{far}}\}$, with
$\delta_{\mathrm{far}}\sim\mathcal{U}\{5,\ldots,32\}$ sampled once per
episode. Closed-loop PASSAGE instead uses $\{0,1,2,3,4,5\}$, while the critic
uses $\{0,1,2,4,8,16,32\}$. Out-of-range indices are clamped at clip
boundaries. Because the planner supplies 25 frames every eight control steps,
no padding is required during normal execution.

The current elevation map
$\mathbf{E}_t\in\mathbb{R}^{3\times31\times61}$ is expressed in the same
torso-centered, yaw-aligned coordinate frame used by the planner, at
$0.05~\mathrm{m}$ horizontal resolution.
The tracker consumes downward depth clipped to $[-3,3]~\mathrm{m}$ directly,
without a validity mask or the planner-specific normalization described above.
During deployment, the most recent map is held between the $20~\mathrm{Hz}$
resampling updates. Map reconstruction and missing-cell handling are detailed in
Appendix~\ref{app:deployment}. The terrain encoder uses three ELU convolutions
with $3\times3$ kernels, stride 2, padding 1, and channels
$3\!\rightarrow\!24\!\rightarrow\!48\!\rightarrow\!96$. Adaptive
$2\times4$ average pooling and a linear $96\!\rightarrow\!256$ projection
produce eight terrain tokens. Tokenizing the six actor reference frames gives
$\mathbf{Z}^{\mathrm{ref}}_t\in\mathbb{R}^{6\times256}$, and fusion is
\begin{equation}
\mathbf{Z}^{E}_t=f_E(\mathbf{E}_t)\in\mathbb{R}^{8\times256},\qquad
\overline{\mathbf{Z}}^{\mathrm{ref}}_t
=[\mathbf{Z}^{\mathrm{ref}}_t;\mathbf{Z}^{E}_t].
\label{eq:terrain_fusion}
\end{equation}
The actor concatenates six reference tokens with the eight terrain tokens,
giving $14\times256$ keys and values for cross-attention. The critic uses seven
reference and eight terrain tokens, giving $15\times256$. Actor and critic
terrain encoders share the architecture but not their parameters.

\subsection{Control and Perceptive Adaptation}
\label{app:tracker_training}

The actor predicts a 29-D residual action. During training, actions are sampled
from a diagonal Gaussian initialized with standard deviation 0.8; deployment
uses the mean. Joint targets and simulated torques are
\begin{equation}
\begin{aligned}
\mathbf{q}^{\mathrm{des}}_t
&=\mathbf{q}^{\mathrm{nom}}+\mathbf{s}_a\odot\mathbf{a}_t,\\
\boldsymbol{\tau}_t
&=\operatorname{clip}\!\left(
\mathbf{K}_p(\mathbf{q}^{\mathrm{cmd}}_t-\mathbf{q}_t)
-\mathbf{K}_d\dot{\mathbf{q}}_t,
-\boldsymbol{\tau}_{\max},\boldsymbol{\tau}_{\max}\right).
\end{aligned}
\label{eq:tracker_action}
\end{equation}
Under joint-encoder-bias randomization,
$\mathbf{q}^{\mathrm{cmd}}_t=\mathbf{q}^{\mathrm{des}}_t-\mathbf{b}_q$.
Torque saturation remains active. The nominal pose uses hip pitch $-0.312$, knee
$0.669$, ankle pitch $-0.363$, shoulder pitch $0.2$, shoulder roll
$+0.2$ (left)/$-0.2$ (right), and elbow $0.6~\mathrm{rad}$, with all other
angles zero. Table~\ref{tab:tracker_control} reports the joint-group parameters.
Each target is held for four $5~\mathrm{ms}$ MuJoCo steps.

\begin{table}[t]
\caption{Residual-action scales, low-level gains, and torque limits.}
\label{tab:tracker_control}
\centering
\scriptsize
\setlength{\tabcolsep}{2.5pt}
\begin{tabularx}{\columnwidth}{@{}Xrrrr@{}}
\toprule
Joint group & $s_a$ (rad) & $K_p$ & $K_d$ & $\tau_{\max}$ ($\mathrm{N\,m}$) \\
\midrule
Hip pitch/roll; knee        & .3507 & 99.10 & 6.31 & 139 \\
Hip yaw; waist yaw          & .5476 & 40.18 & 2.56 & 88 \\
Ankle; waist pitch/roll     & .3070 & 28.50 & 1.81 & 35 \\
Shoulder; elbow; wrist roll & .4386 & 14.25 & .91  & 25 \\
Wrist pitch/yaw             & .0849 & 39.48 & 2.51 & 13.4 \\
\bottomrule
\end{tabularx}
\end{table}

Perceptive adaptation is performed in MJLab/MuJoCo using 90 scenes from the
easy training split, all disjoint from the 150 held-out evaluation scenes.
Each environment samples only references aligned with its assigned scene, and
the three-layer maps are ray-cast online from simulator geometry. The tracker
therefore sees exact simulated geometry during adaptation; reconstruction
error, map latency, and terrain-cell corruption are not simulated. The tracker
is trained only with dataset references and never rolls out the planner. It
remains a single shared policy across all adaptation scenes and motions.

For the first 200 PPO iterations, only the actor terrain encoder and complete
critic are trained; all other actor modules remain frozen. We then unfreeze the
complete actor and continue adaptation for 3{,}800 iterations. The resulting
iteration-4{,}000 checkpoint is fixed across all PASSAGE configurations except
the main-text ``w/o tracker perception'' ablation. Training is distributed
across 32 RTX 4090 GPUs with 2{,}048 environments and 64-step rollouts. We use Adam with an adaptive-KL schedule, actor and critic learning rates of $10^{-5}$ and $5\times10^{-4}$, two update epochs, 16 minibatches per GPU, PPO clip 0.2, desired KL 0.01, $\gamma=0.99$, $\lambda=0.95$, entropy coefficient 0.005, value coefficient 1, and gradient clipping at 1.0.

Here, tracker-side perception augmentation denotes the complete two-stage
procedure that adds terrain conditioning and adapts the tracker. The ablation
restores the base ScaleBFM checkpoint while retaining the same
kinematic-reference interface; it is therefore a stage-level ablation rather
than an isolation of the terrain tokens alone.

\subsection{Rewards and Randomization}
\label{app:tracker_rewards}

The tracker-adaptation objective below is distinct from the planner-side RL
objective in Appendix~\ref{app:planner_rl}.
Let $k_\sigma(e)=\exp(-e/\sigma^2)$. The tracking reward contains pelvis
position and orientation terms with $(w,\sigma)=(0.5,0.3)$ and $(0.5,0.4)$;
mean non-pelvis link position, orientation, linear-velocity, and
angular-velocity terms with $(w,\sigma)=(1,0.3)$, $(1,0.4)$, $(1,1.0)$, and
$(1,3.14)$; and joint-position and joint-velocity terms with
$(w,\sigma)=(0.5,1.25)$ and $(0.25,12)$, respectively. Position and velocity
terms use squared Euclidean errors inside $k_\sigma$, and orientation terms use
squared geodesic errors. Let $c_f\in\{0,1\}$ denote contact of foot $f$; all
sums over $f$ include both feet. Additional rewards are
$0.5\exp(-13\sum_f|z_f-z_f^\star|)$ for foot height,
$0.1\exp(-25\sum_f c_f(1-[\mathbf{R}_f]_{zz}))$ for flat-foot contact, and a
unit survival reward. The penalties are
\begin{equation}
\begin{aligned}
&-0.1\|\mathbf{a}_t-\mathbf{a}_{t-1}\|^2
-10c_{\mathrm{joint}}-0.1c_{\mathrm{self}}-0.1c_{\mathrm{foot}}\\
&\quad-0.75\sum_f c_f\|\mathbf{v}_{f,xy}\|^2
-0.005\sum_{j\in\mathrm{knee}}
\left(\frac{[-\tau_j\dot q_j-5]_+}{50}\right)^2\\
&\quad-0.02\sum_j(\tau_j/\tau_{j,\max})^2
-c_{\mathrm{trk\text{-}contact}},
\end{aligned}
\label{eq:tracker_penalties}
\end{equation}
where $c_{\mathrm{joint}}$ penalizes violations of the inner 90\% joint range,
$c_{\mathrm{self}}$ counts self-contact channels above $10~\mathrm{N}$,
$c_{\mathrm{foot}}$ is the mean reference-contact mismatch of the two feet,
and $c_{\mathrm{trk\text{-}contact}}$ counts robot--obstacle contact channels
above $1~\mathrm{N}$. This force gate is used only by the adaptation reward;
system evaluation follows the contact criterion in the main text. The total
reward is multiplied by the $0.02$-s policy period.
There is no terrain-reconstruction or explicit clearance objective; geometric
conditioning is learned through tracking, survival, and obstacle-contact
returns.

Tracker-adaptation episodes last at most $10~\mathrm{s}$ and terminate when
the pelvis-height error exceeds $0.25~\mathrm{m}$, pelvis-orientation error exceeds
$1~\mathrm{rad}$, ankle or wrist vertical error exceeds $0.25~\mathrm{m}$, or
any reference-link position error exceeds $0.5~\mathrm{m}$. Obstacle contact
alone is not terminal.

At environment construction, we randomize simulated joint-zero offsets by
$\pm0.01~\mathrm{rad}$, foot--ground friction in $[0.6,2.8]$, torso COM by
$\pm(0.025,0.05,0.05)~\mathrm{m}$, hand payload in $[0,1]~\mathrm{kg}$, and
joint-encoder bias by $\pm0.015~\mathrm{rad}$. At reset, PD gains are scaled in
$[0.9,1.1]$; root position, root orientation, joint pose, linear velocity, and
angular velocity are perturbed within
$\pm(0.05,0.05,0.01)~\mathrm{m}$,
$\pm(0.1,0.1,0.2)~\mathrm{rad}$, $\pm0.1~\mathrm{rad}$,
$\pm(0.5,0.5,0.2)~\mathrm{m/s}$, and
$\pm(0.52,0.52,0.78)~\mathrm{rad/s}$, respectively. Velocity perturbations
recur every $1$--$3~\mathrm{s}$. Per-step noise is applied to projected gravity
($\pm0.05$), pelvis angular velocity ($\pm0.2~\mathrm{rad/s}$), joint position
($\pm0.01~\mathrm{rad}$), joint velocity ($\pm0.5~\mathrm{rad/s}$ before
scaling), and reference-pose features ($\pm0.05$).

\section{Onboard Deployment Details}
\label{app:deployment}

\subsection{Multi-Layer Elevation Map Reconstruction}
\label{sec:multilayer_mapping}

Registered point clouds and odometry are fused into the robocentric 3D
occupancy grid maintained by ROG-Map~\cite{ren2023rogmap}. From this grid, we
construct the same torso-centered, yaw-aligned three-layer elevation map used
in training, with 31 lateral and 61 longitudinal samples at
$0.05~\mathrm{m}$ horizontal resolution. For horizontal cell $(u,v)$, let
$\mathcal{V}_{uv}(k)\in\{\mathrm{O},\mathrm{F},\mathrm{U}\}$, denoting
occupied, known-free, and unknown space. With the vertical index increasing
upward, for a column containing occupied voxels we define
\begin{equation}
\begin{aligned}
k^{\mathrm{top}}_{uv}
&=\max\{k\mid \mathcal{V}_{uv}(k)=\mathrm{O}\},\\
\mathcal{K}^{\mathrm{mid}}_{uv}
&=\left\{k\leq k^{\mathrm{top}}_{uv}\ \middle|
\substack{\mathcal{V}_{uv}(k)=\mathrm{O}\\
\mathcal{V}_{uv}(k-1)=\mathrm{F}}\right\},\\
\mathcal{K}^{\mathrm{bot}}_{uv}(k_m)
&=\left\{k<k_m-1\ \middle|
\substack{\mathcal{V}_{uv}(k)=\mathrm{O}\\
\mathcal{V}_{uv}(k+1)=\mathrm{F}}\right\}.
\end{aligned}
\label{eq:three_layer_extraction}
\end{equation}
The top index gives the highest occupied surface. If
$\mathcal{K}^{\mathrm{mid}}_{uv}$ is nonempty, set
$k^{\mathrm{mid}}_{uv}=\max\mathcal{K}^{\mathrm{mid}}_{uv}$ and evaluate
$\mathcal{K}^{\mathrm{bot}}_{uv}(k^{\mathrm{mid}}_{uv})$; if the latter is
also nonempty, set $k^{\mathrm{bot}}_{uv}$ to its maximum. Let $z_+(k)$ and
$z_-(k)$ denote the upper and lower faces of voxel $k$. When
$z_-(k^{\mathrm{mid}}_{uv})-z_+(k^{\mathrm{bot}}_{uv})\geq0.05~\mathrm{m}$,
the overhang is encoded by
$z^{\mathrm{top}}_{uv}=z_+(k^{\mathrm{top}}_{uv})$,
$z^{\mathrm{mid}}_{uv}=z_-(k^{\mathrm{mid}}_{uv})$, and
$z^{\mathrm{bot}}_{uv}=z_+(k^{\mathrm{bot}}_{uv})$. The latter two surfaces
are therefore an obstacle underside certified by observed free space beneath
it and the supporting surface below that clearance. If either transition is
absent or the clearance is smaller, $z_+(k^{\mathrm{top}}_{uv})$ is repeated
across the three channels. Unknown voxels never satisfy a known-free predicate
and are therefore never used to certify an underside or clearance. Missing
layer values are completed only in the elevation cache; completion neither
changes occupancy labels nor certifies unknown space as free.

To suppress sparse-LiDAR artifacts, within a detected-overhang mask and its
one-cell dilation we compare each support candidate with the 75th percentile
of valid supports in a $5\times5$ neighborhood. A candidate more than
$0.4~\mathrm{m}$ above this reference is replaced when at least four neighbors
agree within two vertical voxels. Queried columns without a valid support are
completed by iteratively propagating the minimum support height among their
eight-connected neighbors; for a column with no occupied surface, the
completed support is repeated across the three channels. Both operations
modify only the elevation cache, not the underlying occupancy grid.

The resulting world-frame surface heights $z^{\ell}_{uv}$ are converted to
torso-relative downward depths,
\begin{equation}
E_i^{\ell}(u,v)=
\operatorname{clip}\!\left(
z_i^{\mathrm{torso}}-z^{\ell}_{uv},
-3~\mathrm{m},3~\mathrm{m}\right).
\label{eq:three_layer_observation}
\end{equation}
Here, $\ell\in\{\mathrm{top},\mathrm{mid},\mathrm{bot}\}$.
The tracker consumes these clipped metric depths, whereas the planner applies
the normalization in Appendix~\ref{app:planner_architecture}. Surface
extraction is updated with accepted point clouds, and the cached surfaces are
resampled in the torso frame at $20~\mathrm{Hz}$ using the latest odometry.

\subsection{Onboard Runtime Details}
\label{sec:realtime_integration}

The complete stack runs onboard a Unitree G1 equipped with a Jetson AGX Orin
and a Manifold Tech Odin module, without a prebuilt map or offboard computation.
The planner is exported to ONNX and executed in FP16 through the ONNX Runtime
TensorRT provider in a dedicated OS process. A new plan is requested every
eight $50~\mathrm{Hz}$ control steps, giving the reported
$6.25~\mathrm{Hz}$ planning rate. Requests and results are exchanged through a
multiprocessing pipe, while the tracker executes on the onboard CPU in the
$50~\mathrm{Hz}$ control thread. The latest joint target is transferred
through POSIX shared memory to an out-of-process C++ bridge and republished to
the robot at $500~\mathrm{Hz}$.

\clearpage
\balance
\section{Zhang et al.-Style Diffusion--Tracking Baseline}
\label{app:zhang_baseline}

We provide implementation details for the Zhang et al.-style
diffusion--tracking pipeline evaluated in the main text. This is our
adaptation rather than an evaluation or reproduction of the authors'
original system. We retain the reported 25-frame horizon, two history
frames, two-step online denoising, and closed-loop tracker adaptation
with the motion generator frozen. The diffusion planner and a
task-specific perceptive tracker are trained from scratch using the
same unaugmented, nested 12-h subset. Because of simulator
scene-loading constraints, this implementation uses a tracker
specialized to that subset rather than a general tracker.

\textbf{Diffusion planner.}
For compatibility with our robot and benchmark, the planner uses
PASSAGE's 65-D motion state, 5-D torso-yaw-frame destination, and
$3\times31\times61$ elevation map. Its terrain encoder contains three
stride-1 $3\times3$ Conv--GN--SiLU layers
($3\!\rightarrow\!32\!\rightarrow\!64\!\rightarrow\!64$), replicate
padding to $33\times63$, and a stride-3 projection to 256-D, yielding
231 terrain tokens. The denoiser comprises eight pre-norm Transformer
blocks with width 256, eight attention heads, FFN width 1024, GELU,
dropout 0.05, and global bidirectional attention. We use discrete VP
diffusion with $T=21$, a squared-cosine schedule ($s=0.008$),
$\beta$ values clipped to $[10^{-4},0.9999]$, and direct clean-motion
prediction. The training objective combines clean-motion MSE with
velocity and jerk losses weighted by 0.05 and 0.02, respectively; the
jerk loss uses a $1{,}000~\mathrm{m/s^3}$ hinge threshold.

The planner is trained for $4{,}000$ epochs using AdamW, batch size 64
per rank, learning rate $10^{-4}$, $(\beta_1,\beta_2)=(0.9,0.95)$,
$\epsilon=10^{-8}$, weight decay $10^{-4}$, gradient-norm clip 10,
and seed 42, without warmup, EMA, or learning-rate scheduling.
At inference, deterministic DDIM evaluates the denoiser at $k=20$ and
$k=0$ with $\eta=0$. At each step, history-conditioned and
history-masked predictions are blended with equal weight, requiring
four forward passes per replan. Fresh Gaussian noise is sampled at
every replan, and the first eight predicted frames are executed,
giving $6.25~\mathrm{Hz}$ planning under $50~\mathrm{Hz}$ tracking.

\textbf{Tracker pre-training.}
Following the dataset-only tracker-training stage of Zhang et al., we
train a task-specific perceptive tracker from scratch on the same
unaugmented 12-h motion--scene subset. Zhang et al. use an RL-based
whole-body reference tracker; the ScaleBFM-compatible backbone used
here is our implementation choice for compatibility with our 29-DoF
robot and benchmark. It contains four Transformer blocks with width
256, four attention heads, FFN width 256, and a
$3\!\rightarrow\!24\!\rightarrow\!48\!\rightarrow\!96$ terrain CNN
that produces eight terrain tokens. The policy outputs 29 residual
joint-position commands. All model parameters, including the terrain
encoder, are randomly initialized. During this stage, the tracker is
trained only with dataset references and does not receive
planner-generated rollouts.

Tracker pre-training uses separate Adam optimizers with actor and
critic learning rates of $10^{-5}$ and $5\times10^{-4}$, desired KL
0.01, PPO clip 0.2, $\gamma=0.99$, $\lambda=0.95$, entropy coefficient
0.005, two update epochs, 16 minibatches, and gradient-norm clip 1.0.
Each iteration collects 64 simulator steps per environment, and
training runs for $14{,}000$ iterations.

\textbf{Closed-loop tracker adaptation.}
After dataset-only tracker pre-training, the diffusion planner remains
frozen in evaluation mode, while the tracker actor, terrain encoder,
and critic are optimized using planner-generated references
conditioned on executed robot-state history. This is the only stage in
which the tracker receives planner-generated rollouts. PPO augments
the tracker imitation and regularization rewards with a
destination-heading reward and a robot--obstacle contact penalty.
Each iteration collects 24 simulator steps per environment using
$1{,}024$ environments per rank. Adaptation continues for
$20{,}000$ iterations with the same PPO coefficients and learning
rates as tracker pre-training, and the final checkpoint is evaluated.

As stated in the main text, differences in tracker architecture,
training, and closed-loop refinement mean that this experiment
compares complete pipelines rather than isolating flow matching from
diffusion.

\end{document}